\documentclass[]{spie}  

\usepackage{amsmath,amsfonts,amssymb}
\usepackage{graphicx}
\usepackage{cite} 
\usepackage{times}
\usepackage{epsfig}
\usepackage{amsmath}
\usepackage{nccmath}
\usepackage{amssymb}
\usepackage{mwe}
\usepackage{acro}
\usepackage{amssymb}
\usepackage{xcolor,colortbl}
\usepackage{tabularx}
\usepackage{relsize}
\usepackage{pifont}
\usepackage{booktabs} 
\usepackage{multirow}
\usepackage{multicol}
\usepackage{adjustbox}
\usepackage{float}
\usepackage{graphicx}
\usepackage{makecell}
\usepackage{tabu}
\usepackage[colorlinks=true, allcolors=blue]{hyperref}
\usepackage[capitalize]{cleveref}

\usepackage{algorithm}
\usepackage{algorithmic}

\title{Enhanced Feature-based Image Stitching for Endoscopic Videos in Pediatric Eosinophilic Esophagitis}

\author[a]{Juming Xiong}
\author[b]{Muyang Li}
\author[b]{Ruining Deng}
\author[b]{Tianyuan Yao}
\author[a]{Shunxing Bao}
\author[c]{Regina N Tyree}
\author[c]{Girish Hiremath}
\author[a,b]{Yuankai Huo}

\affil[a]{Department of Electrical and Computer Engineering, Vanderbilt University, Nashville, TN, USA}
\affil[b]{Department of Computer Science, Vanderbilt University, Nashville, TN, USA}
\affil[c]{Division of Pediatric Gastroenterology, Hepatology, and Nutrition, Vanderbilt University Medical Center, Nashville, TN, USA}

\authorinfo{Corresponding author: Yuankai Huo: E-mail: yuankai.huo@vanderbilt.edu}

\begin{document} 
\maketitle

\begin{abstract}
Video endoscopy represents a major advance in the investigation of gastrointestinal diseases. Reviewing endoscopy videos often involves frequent adjustments and reorientations to reconstruct a comprehensive view of the examined area, a process that is not only time-consuming but also prone to errors. Image stitching techniques offer a solution by enabling a continuous and complete visualization of the examined area. However, endoscopic images, particularly those of the esophagus, present unique challenges. The smooth surface, lack of distinct feature points, and non-horizontal orientation complicate the stitching process, rendering traditional feature-based methods often ineffective for these types of images. To solve this problem, we propose a novel preprocessing pipeline designed to enhance endoscopic image stitching through advanced computational techniques. Our approach converts endoscopic video data into continuous 2D images by following four key steps: (1) keyframe selection, (2) image rotation adjustment to correct distortions, (3) surface unwrapping using polar coordinate transformation to generate a flat image, and (4) feature point matching enhanced by Adaptive Histogram Equalization for improved feature detection. We evaluate stitching quality through the assessment of valid feature point match pairs. Experiments conducted on 20 pediatric endoscopy videos demonstrate that our method significantly improves image alignment and stitching quality compared to traditional techniques, laying a robust foundation for more effective panoramic image creation.

\end{abstract}

\keywords{Eosinophilic esophagitis, Endoscopy, Image stitching}

\section{INTRODUCTION}
\label{sec:intro}  
Endoscopic procedures are integral to the diagnosis and treatment of gastrointestinal (GI) discoders, offering direct visualization of the digestive tract~\cite{Gulati2020-qq,6991521}. Esophageal endoscopy plays an important role in detecting and monitoring conditions such as Eosinophilic esophagitis (EoE)\cite{article1}, and Barrett's esophagus\cite{fitzgerald2014british}. Achieving high-quality visualizations during endoscopic examinations is essential for accurate diagnosis and effective patient management.

\begin{figure*}[t]
\begin{center}
\includegraphics[width=1\linewidth]{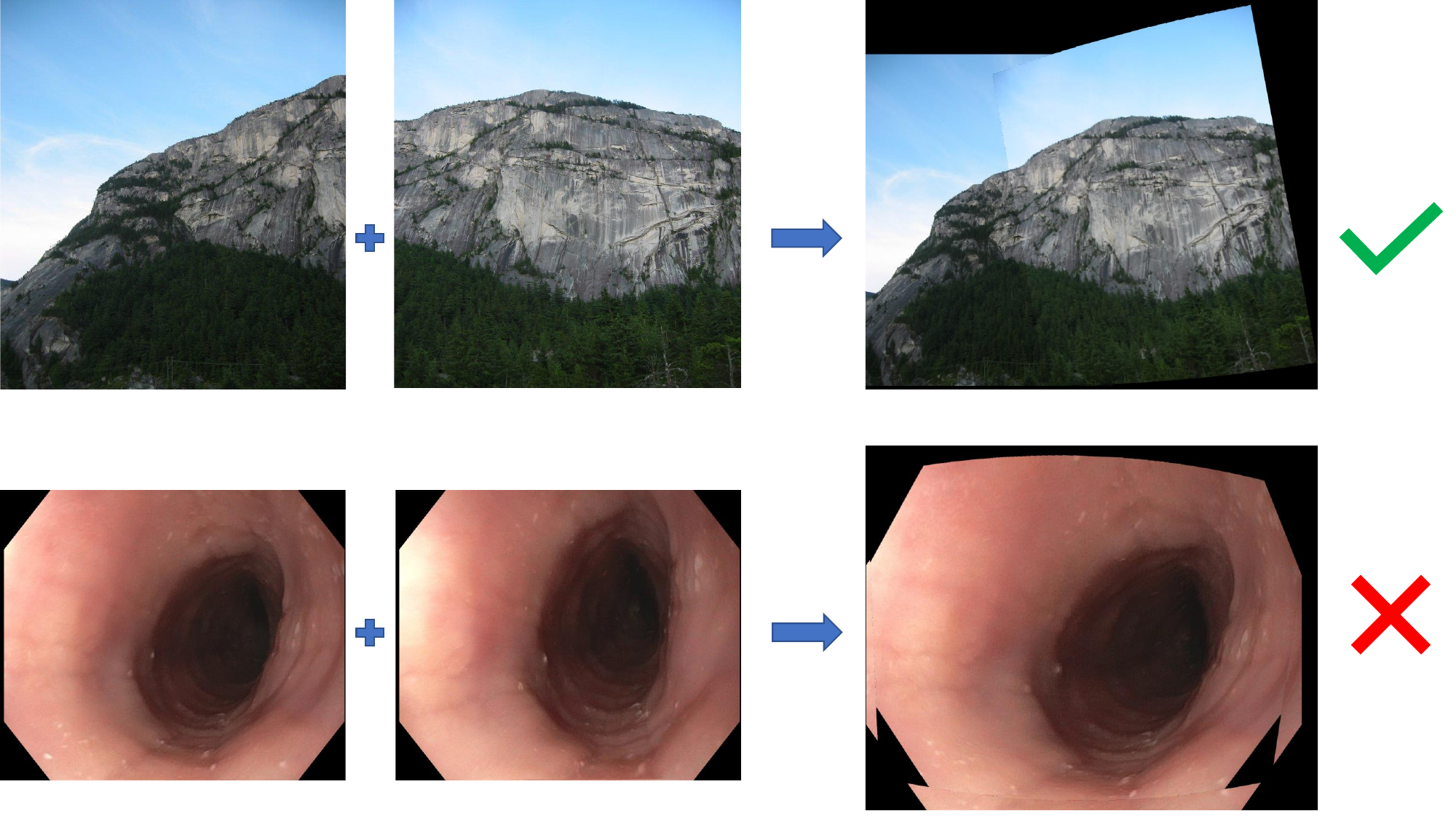}
\end{center}
\caption{The upper panel shows the general image stitching result and it can be successfully stitched. The lower panel shows the failed endoscopic image stitching result by using the traditional feature-based image stitching method.}
\label{fig:overview}
\end{figure*}

Despite the essential role of endoscopy in diagnosing esophageal conditions, the process of reviewing endoscopic videos remains a time-consuming issue. Clinicians often struggle with the frequent need to reorient and interpret fragmented views to capture a comprehensive understanding of the examined areas. A major challenge in this context is the effective stitching of endoscopic images to produce continuous, panoramic views. Traditional image stitching techniques, which typically rely on the presence of abundant feature points and horizontal alignment of images, are often inadequate for esophageal endoscopy. The smooth texture of the esophageal lining and the non-horizontal orientation of the captured images introduce considerable difficulties, making it challenging to achieve reliable and accurate stitched images as shown in Figure~\ref{fig:overview}. These limitations highlight the need for innovative solutions to improve the efficiency and accuracy of esophageal endoscopy.

In this paper, we propose a novel pipeline designed to enhance endoscopic image stitching. Our approach is particularly aimed at converting endoscopic video data into continuous 2D images, thereby facilitating better visualization and diagnosis. The pipeline comprises four key steps: keyframe selection, image rotation adjustment, surface unwrapping via polar coordinate transformation, and feature point matching using Adaptive Histogram Equalization (AHE). This pipeline facilitates the transformation of esophageal video data into 2D images and establishes a solid foundation for achieving comprehensive panoramic esophageal image stitching.

\section{Method}
The proposed pipeline includes four steps: (1) Keyframe selection, (2) Image rotation adjustment, (3) Unwrapping, and (4) Feature point matching

\begin{figure*}[t]
\begin{center}
\includegraphics[width=1\linewidth]{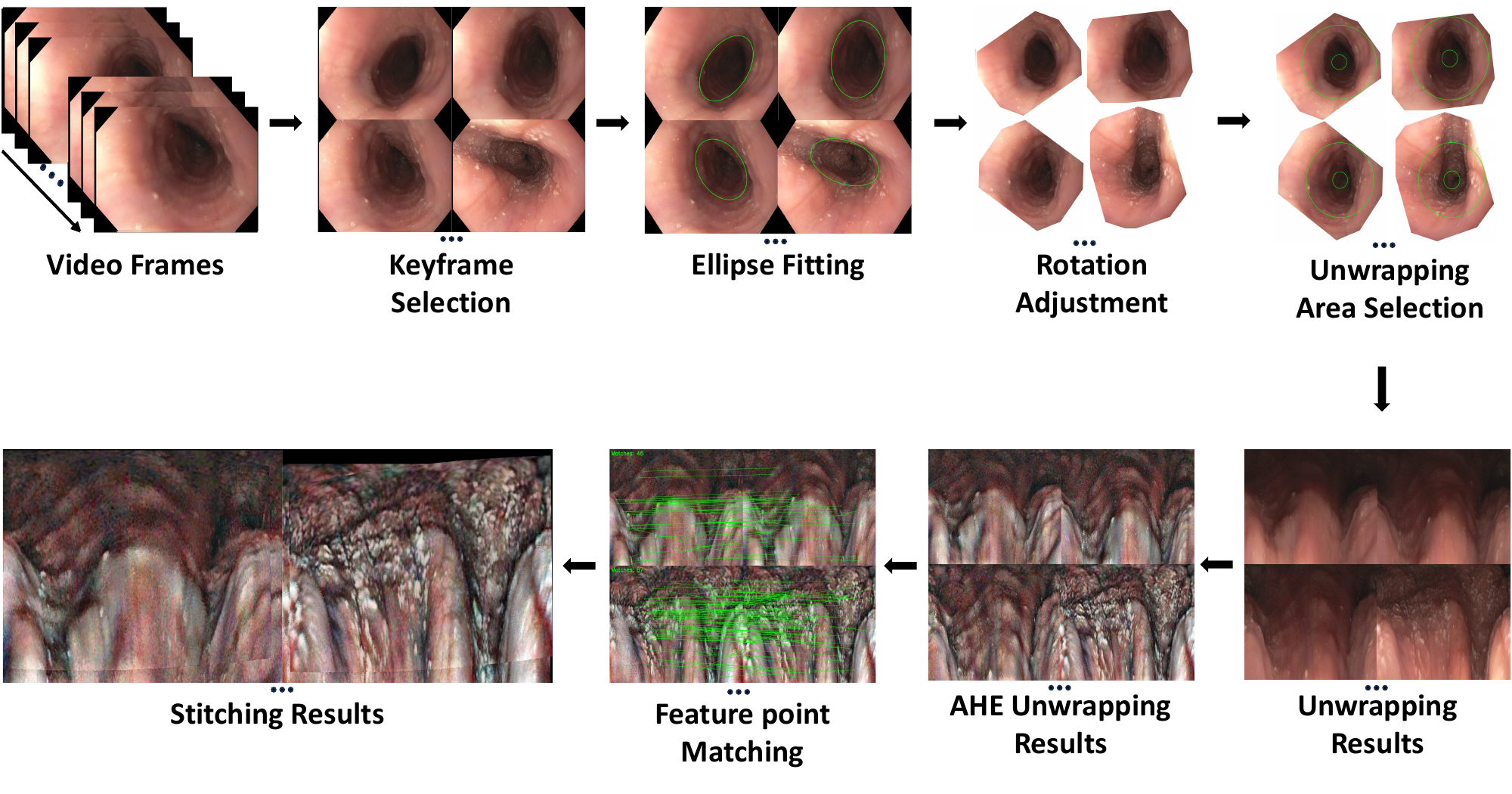}
\end{center}
\caption{This figure illustrates an enhanced endoscopic image preprocessing pipeline designed for stitching endoscopic images. The process begins with the input endoscopic images, which undergo ellipse fitting to identify and mark regions of interest. Following ellipse fitting, the images are subjected to rotation adjustment to standardize their orientation. Specific unwrapping areas are then selected to facilitate detailed analysis. The unwrapping result provides a flattened view of the mucosal surface. Then, AHE is applied to the unwrapped images, enhancing features and improving the quality of the images for feature point matching. Finally, this process produces the stitching results. }
\label{fig:Method}
\end{figure*}

\subsection{Keyframe Selection}
To address the challenges posed by motion blur and occlusion in endoscopic videos, we implement a robust keyframe selection method. Given the tendency for frames to be particularly blurry during camera movement and the common issue of occlusion at the beginning and end of the videos, we drop the first 3 seconds and the last 3 seconds of each video. From the remaining footage, we apply Uniform Sampling to extract one frame every 5 frames. 

\subsection{Image rotation adjustment}
Due to the uncontrollable rotation of the capsule camera when navigating the esophagus, the resulting images exhibit inconsistencies upon unwrapping, which adversely affects stitching results. To mitigate this issue, it is essential to adjust the image orientation to counteract the effects of rotation.

Our methodology starts with identifying the deep area of the image, which is typically characterized by lower pixel values and often exhibits an elliptical shape. To accurately delineate this area, we apply a pixel value threshold to extract the contour of the region below this threshold. We then fit an ellipse to this contour, which allows us to ascertain the orientation of the deep area. By utilizing the angle of the fitted ellipse, we adjust the image to compensate for the rotational distortions and improve the stitching outcome.

\subsection{Unwrapping}
After adjusting the angle, we identify the deepest point of the image, typically characterized by the smallest pixel value. To ensure robustness against noise, we set a threshold and compute the average position of all points below this threshold, thus mitigating the influence of outliers. This averaged deepest point is used as the center to define two concentric circles~\cite{app9163437}. The larger concentric circle is as large as possible without exceeding the image boundary, while the smaller concentric circle has a radius minimized under the set threshold.

We then flatten the image using the annular area between these concentric circles as the effective region. Polar coordinate transformation is employed to convert this annular region into a flat image, and interpolation is used to fill in any missing parts.

The unwrapping process can be described mathematically. Let \((r, \theta)\) be the polar coordinates of a point in the annular region, where \(r\) is the radius and \(\theta\) is the angle. These coordinates can be related to the Cartesian coordinates \((x, y)\) of the image as follows:

\begin{align}
x &= x_0 + r \cos(\theta) \\
y &= y_0 + r \sin(\theta)
\end{align}

where \((x_0, y_0)\) is the center of the concentric circles.

To perform the polar coordinate transformation, we map the coordinates from the polar system to a rectangular grid, effectively "unwrapping" the annular region:

\begin{align}
u &= r \\
v &= \theta
\end{align}

Here, \(u\) and \(v\) are the coordinates in the transformed image. Interpolation is applied to determine the pixel values in the unwrapped image, ensuring smooth transitions and filling any gaps:

\begin{equation}
I'(u, v) = I(x, y)
\end{equation}

where \(I(x, y)\) is the original image intensity at \((x, y)\) and \(I'(u, v)\) is the intensity in the transformed image at \((u, v)\). This transformation effectively addresses the rotational inconsistencies and enhances the stitching quality.

\subsection{Feature point matching}

Due to the relatively smooth texture of the esophageal interior, there are fewer distinguishable feature points compared to traditional or other medical images. This scarcity of features often results in image stitching failures or suboptimal results.

To address this issue, we apply Adaptive Histogram Equalization (AHE)~\cite{PIZER1987355} to each endoscopic image. AHE is a technique that improves the local contrast of different regions within an image, thereby enhancing its features. By applying AHE, we aim to increase the number of detectable feature points, which are essential for successful image stitching.

Following the application of AHE, we utilize the Scale-Invariant Feature Transform (SIFT) algorithm to identify and match feature points between images. SIFT is a robust technique that detects and describes local features, generating feature points that are invariant to changes in scale and rotation~\cite{inbook}. To enhance the quality of the initial matches, we first apply a filtering criterion: we compare the distances between each pair of feature points and retain only those pairs where the distance of the match is less than 0.75 times the distance of the second-best match. This filtering step ensures that only the most reliable matches are selected for further processing.

Subsequently, we employ the Random Sample Consensus (RANSAC) algorithm to refine these filtered matches. RANSAC estimates the parameters of a mathematical model from the filtered data, which may still contain outliers~\cite{FISCHLER1987726}. By focusing on the most reliable matches, RANSAC effectively reduces the impact of any remaining outliers, ensuring a more accurate and robust image stitching process. This combined methodology not only enhances the quality of the feature point matches but also considerably improves the overall stitching accuracy and reliability.

\section{Data and Experiments}

\subsection{Data}
\begin{figure*}[t]
\begin{center}
\includegraphics[width=1\linewidth]{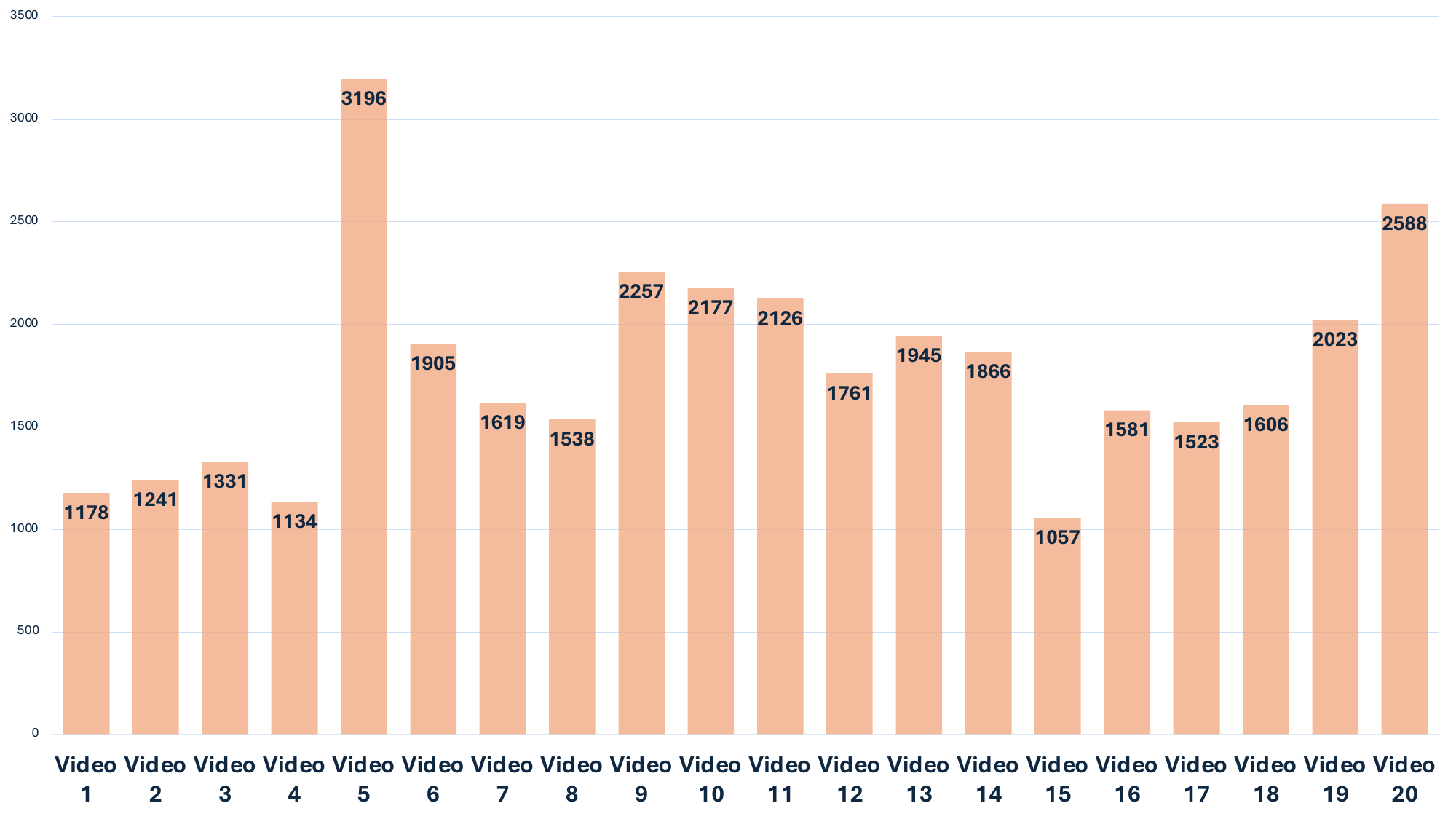}
\end{center}
\caption{This figure illustrates the total number of videos analyzed and the corresponding number of frames extracted from these videos.}
\label{fig:data}
\end{figure*}

In this study, 20 videos, comprising 35,652 frames as detailed in Figure~\ref{fig:data}, were collected from five children aged 6-18 diagnosed with EoEsinophilic Esophagitis (EoE) at Monroe Carrell Jr. Children’s Hospital at Vanderbilt University Medical Center (VUMC). The study protocol, which did not involve any procedures beyond routine care, received approval from the Institutional Review Board at Vanderbilt University Medical Center. Participants were enrolled after obtaining the necessary consent from their caregivers and assent from the children.

\subsection{Experiment}

\subsubsection{Evaluation metric}
In the field of image stitching, the absence of ground truth data poses substantial challenges for directly evaluating the accuracy of the stitching results. To address this issue, the number of valid feature point match pairs is employed as an evaluation metric. This metric involves detecting and matching feature points between overlapping images using algorithms such as the SIFT and validating these matches with techniques like RANSAC. The number of valid feature point matches serves as an indicator of the robustness and precision of the feature detection and matching process. A higher count of valid feature point matches generally correlates with better alignment and overlap between the images, leading to more reliable and accurate stitching outcomes. Thus, in the absence of ground truth, the number of valid feature point match pairs provides an indirect yet effective measure of stitching quality.

\begin{figure*}[t]
\begin{center}
\includegraphics[width=1.0\linewidth]{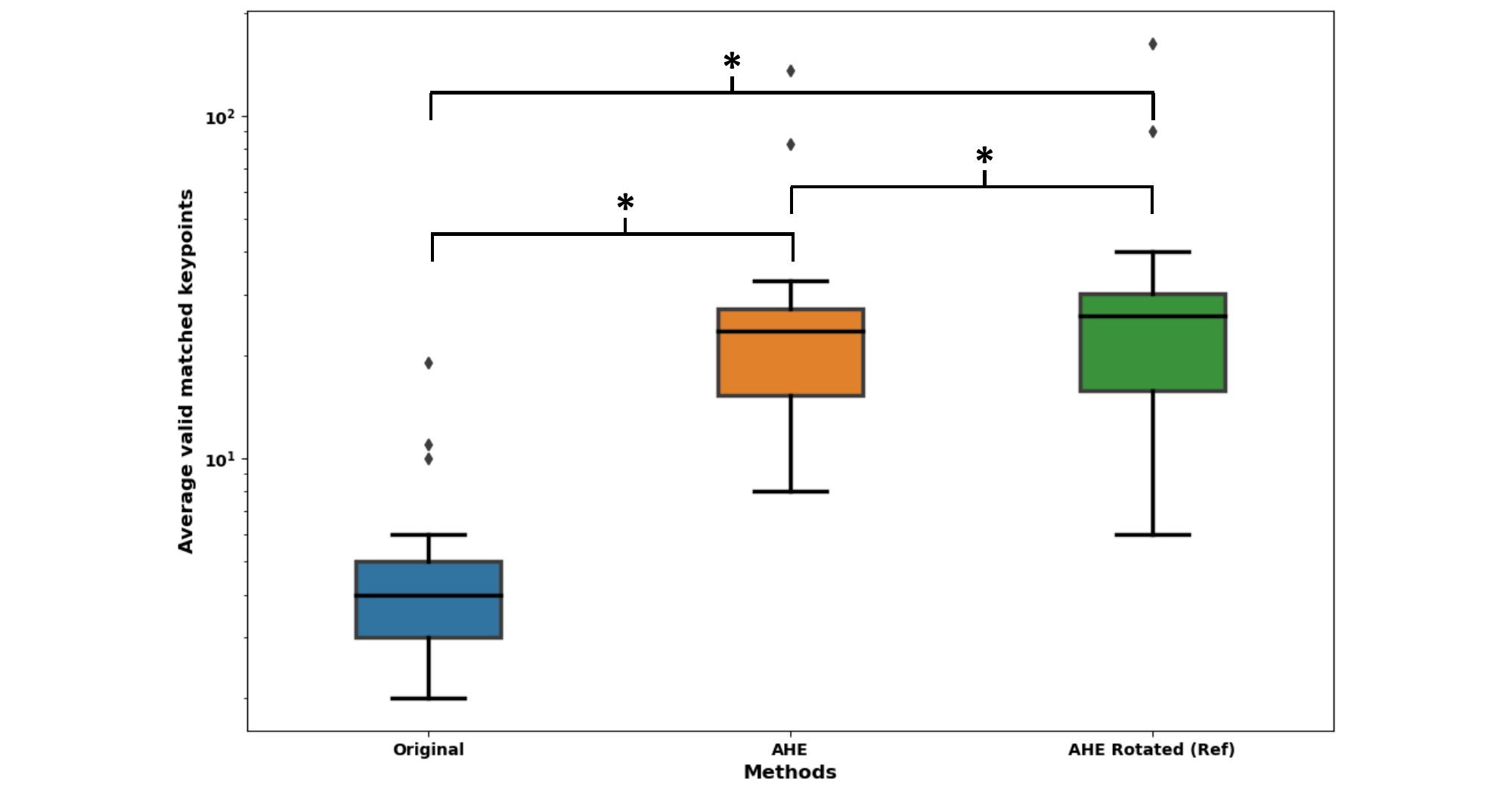}
\end{center}
\caption{This figure presents a boxplot illustrating the number of valid matched feature points for keyframe selection across 20 videos using three different methods. The Original unwrapped images show the lowest number of valid matched feature points. In contrast, applying the AHE method to the unwrapped images markedly increases the number of feature points. The highest number of valid matched feature points is observed with the AHE Rotated unwrapped images, where both AHE and rotation methods are applied. The Wilcoxon signed-rank test is performed with AHE Rotated as the reference (“Ref”) method, to compare with other methods. “*” represents the significant (p $<$ 0.05) differences}
\label{fig:box}
\end{figure*}

\begin{figure*}[t]
\begin{center}
\includegraphics[width=0.8\linewidth]{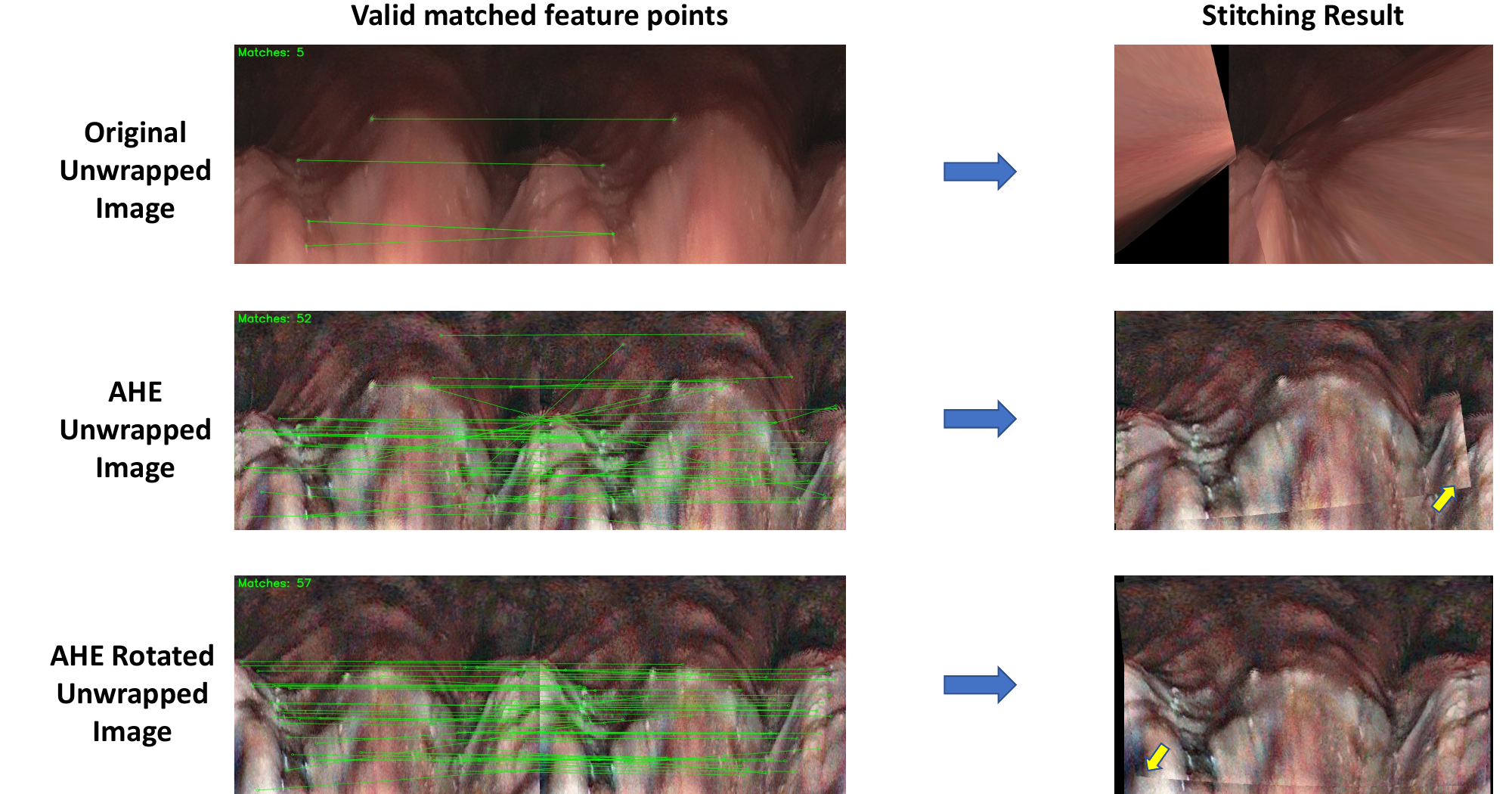}
\end{center}
\caption{Result comparison among different image processing stages. The original unwrapped image shows only 5 valid matched feature points, resulting in a failed stitching outcome. The AHE unwrapped image, with 52 valid matched feature points, demonstrates improved stitching but still exhibits visible gaps between the images As shown by the yellow arrow. The AHE rotated unwrapped image, which has 57 valid matched feature points, achieves the best stitching result, with more seamless alignment as indicated by the yellow arrow.}
\label{fig:visualize}
\end{figure*}

\section{Results}
The average number of valid feature point matching pairs for each selected video frame was evaluated under three methods: Original (original unwrapped image), AHE (AHE-applied unwrapped image), and AHE Rotated (AHE and rotation methods applied unwrapped image). The results indicate that applying AHE and rotation generally enhances the number of valid feature point matches compared to using the original image or AHE alone.

The AHE Rotated method yielded the highest average number of valid feature point matching pairs, demonstrating its superior effectiveness in enhancing image clarity and feature matching accuracy. The AHE Rotated method yielded the highest average number of valid feature point matching pairs, demonstrating its superior effectiveness in enhancing image clarity and feature matching accuracy. Both AHE and AHE Rotated methods significantly improved the number of valid matched feature points compared to the Original method. The Original method shows the lowest number of valid matched feature points, with a narrower range indicating less variability. The AHE Rotated method shows a significant improvement with a higher median value compared to the AHE method, although its minimum value is lower and it is shown in Fig~\ref{fig:box}.

Fig~\ref{fig:visualize} compares image stitching results across different processing stages, focusing on valid matched feature points and stitching quality. The Original Unwrapped Image has only 5 valid feature points, leading to failed stitching result. The AHE Unwrapped Image improves with 52 feature points, but shows large gaps. The AHE Rotated Unwrapped Image achieves 57 feature points, resulting in the best alignment and minimal gaps.

\section{Conclusion}
We introduced a novel pipeline to enhance seamless panoramic endoscopic image stitching, which integrates keyframe selection, rotation adjustment, image unwrapping, and AHE feature enhancement. Our approach begins with keyframe selection, where we discard the first and last 3 seconds of each video and use Uniform Sampling to address issues of motion blur and occlusion, ensuring that the keyframes are clear and representative. Next, we apply rotation adjustment to standardize image orientation, which is crucial for accurate alignment. This is followed by image unwrapping, which transforms the endoscopic images into a planar view, facilitating precise feature matching. Finally, AHE is applied to enhance local contrast and improve feature visibility in the unwrapped images.

The experimental results demonstrate that this comprehensive pipeline increases the number of detectable and matchable feature points, leading to improved image alignment and stitching quality. This method lays a robust foundation for panoramic image stitching in the esophagus, paving the way for more precise subsequent endoscopic image analysis and enhanced diagnostic accuracy..
\newpage

\section{ACKNOWLEDGMENTS}       
This research was supported by NIH R01DK135597(Huo), DoD HT9425-23-1-0003(HCY), NIH NIDDK DK56942(ABF). This work was also supported by Vanderbilt Seed Success Grant, Vanderbilt Discovery Grant, and VISE Seed Grant. This project was supported by The Leona M. and Harry B. Helmsley Charitable Trust grant G-1903-03793 and G-2103-05128. This research was also supported by NIH grants R01EB033385, R01DK132338, REB017230, R01MH125931, and NSF 2040462. We extend gratitude to NVIDIA for their support by means of the NVIDIA hardware grant.

\bibliography{main} 
\bibliographystyle{spiebib} 

\end{document}